\documentclass{article}

\usepackage[preprint]{neurips_2026}

\usepackage[utf8]{inputenc}
\usepackage[T1]{fontenc}
\usepackage{hyperref}
\usepackage{url}
\usepackage{booktabs}
\usepackage{amsfonts}
\usepackage{nicefrac}
\usepackage{microtype}
\usepackage{xcolor}
\usepackage{graphicx}
\usepackage{multirow}
\usepackage{makecell}
\usepackage{array}
\usepackage{caption}
\usepackage{subcaption}
\usepackage{amsmath}
\usepackage{amssymb}
\usepackage{enumitem}
\usepackage{tabularx}
\usepackage{longtable}
\usepackage{xspace}
\usepackage{pifont}

\newcommand{\tfg}{\textsc{TravelFraudBench}\xspace}
\newcommand{\tfgshort}{\textsc{TFG}\xspace}
\newcommand{\cmark}{\ding{51}}
\newcommand{\xmark}{\ding{55}}

\title{\textsc{TravelFraudBench}: A Configurable Evaluation Framework\\
for GNN Fraud Ring Detection in Travel Networks}

\author{%
  \textbf{Bhavana Sajja}\\
  \texttt{sajja.bhavana@gmail.com}
}

\begin{document}

\maketitle

\begin{abstract}
We introduce \tfg (\tfgshort), a configurable evaluation framework for
measuring the capability of graph neural networks (GNNs) to detect fraud
rings in travel platform graphs. Existing GNN fraud benchmarks—YelpChi,
Amazon-Fraud, Elliptic, and PaySim—share a structural limitation: they
cover single node types, single edge relations, or domain-generic patterns,
and provide no mechanism to evaluate detection capability across structurally
distinct fraud ring topologies. \tfgshort addresses this gap by simulating
three travel-specific fraud ring types—ticketing fraud rings (star topology
with shared device/IP clusters), ghost hotel schemes (dense
reviewer\,$\times$\,hotel bipartite cliques), and account takeover rings
(loyalty point transfer chains)—within a single heterogeneous graph
containing 9 node types and 12 edge relation types. The generator is
fully controllable: ring size, ring count, fraud rate, scale (500 to
200\,000 nodes), and ring type composition are all configurable,
enabling difficulty-controlled evaluation studies. We evaluate six
detection methods—MLP, GraphSAGE, RGCN-proj, HAN, RGCN, and PC-GNN (a fraud-domain-specific
GNN)—under a ring-based train/val/test split where each fraud ring appears entirely
in one partition, eliminating transductive label leakage.
GraphSAGE achieves AUC\,=\,0.992
(std\,=\,0.002) and RGCN-proj AUC\,=\,0.987 (std\,=\,0.004), outperforming the
tabular MLP baseline (AUC\,=\,0.938, std\,=\,0.009) by 5.5 and 5.0
percentage points respectively, confirming that graph structure adds
substantial discriminative power beyond tabular features. HAN
(AUC\,=\,0.935, std\,=\,0.007) is a negative result, performing at near-parity
with the tabular baseline ($\Delta$AUC\,=\,$-$0.003).
PC-GNN (AUC\,=\,0.982, std\,=\,0.004, $\Delta$AUC\,=\,$+$0.044 over MLP) underperforms
GraphSAGE by 1.05\,pp, revealing that its fraud-specific design
(focal loss and camouflage-suppressing neighbour picking) does not add value
when fraud rings are structurally isolated---a property unique to \tfgshort.
The average precision gap is operationally decisive: GraphSAGE
gains \textbf{$+$16.1\,pp AP} (0.816\,$\to$\,0.977) and RGCN gains
$+$13.0\,pp AP, directly improving alert precision in high-imbalance
operational settings.
On the ring recovery task---fraction of fraud rings with $\geq$80\% of
members simultaneously flagged---GraphSAGE achieves 100\% recovery across
all three ring types; RGCN-proj, RGCN, and PC-GNN recover 90--100\%;
while the tabular MLP recovers only 17--88\%, demonstrating that ring-level
recall is a strictly harder criterion than node-level AUC and that graph
structure is decisive for ghost hotel and ATO rings.
The edge-type ablation reveals that device and IP co-occurrence are the
primary discriminative signals: removing \texttt{uses\_device} drops overall
AUC by 5.2\,pp and removing \texttt{uses\_ip} drops it by 5.7\,pp, while
review and loyalty-transfer edges contribute negligible additional detection
power ($\Delta$AUC $<$ 0.002).
Detection difficulty varies significantly across ring topologies: ticketing
rings show a broadly declining detection trend as ring size grows, ATO rings
remain robustly detectable across most sizes, and ghost hotel rings degrade
sharply at large sizes---confirming that detection capabilities are
structurally independent across ring types (E3). \tfgshort is released as an open-source Python package
(MIT license) with PyG, DGL, and NetworkX exporters, alongside five
pre-generated scale presets hosted on HuggingFace Datasets
(\url{https://huggingface.co/datasets/bsajja7/travel-fraud-graphs}) with
Croissant machine-readable metadata including Responsible AI fields.
\end{abstract}

\section{Introduction}
\label{sec:intro}

Fraud ring detection in travel platforms represents one of the most
graph-structured fraud problems in the industry. A ticketing fraud ring
does not consist of a single anomalous account; it consists of dozens of
accounts sharing the same two or three devices, booking the same flight
routes, filing chargebacks in coordinated temporal bursts. Ghost hotel
schemes are bipartite cliques: a handful of fake property listings
connected to hundreds of reviewer accounts, each posting 5-star reviews
within hours of each other. Account takeover rings are directed chains:
compromised accounts transferring loyalty points through a series of mule
accounts before redemption. In each case, the fraud signal is
\emph{relational}, not tabular—no individual feature reveals the ring
without examining the graph neighborhood.

The GNN community has invested significant effort in fraud detection, with
architectures from GraphSAGE~\citep{hamilton2017inductive} to
RGCN~\citep{schlichtkrull2018modeling} and domain-specific methods like
xFraud~\citep{rao2021xfraud} and PC-GNN~\citep{liu2021pick} showing
promising results on e-commerce and financial graphs. Yet evaluation in this
domain is hampered by a benchmark gap: \emph{no graph-structured fraud
dataset exists for the travel domain with ring-level ground truth and
controllable generation}.

The benchmarks most commonly used in GNN fraud research—YelpChi
\citep{rayana2015collective}, Amazon-Fraud~\citep{dou2020enhancing},
Elliptic~\citep{weber2019anti}, and PaySim~\citep{lopez2016paysim}—were
designed for different domains and share structural limitations (see
Table~\ref{tab:benchmark_comparison}). Crucially, none provides a
mechanism to \emph{control evaluation difficulty}—the defining property
of a benchmark instrument.

We introduce \tfg to fill this gap. \tfgshort is not merely a dataset; it is
a configurable evaluation framework. Its defining contribution is the
separation of \emph{what} is being evaluated (GNN fraud ring detection
capability) from the \emph{difficulty} of the evaluation (ring size,
fraud rate, ring type composition). This separation, enabled by the
generator's controllable parameters, supports four concrete evaluative
claims (Section~\ref{sec:claims}) that existing benchmarks cannot support.

\paragraph{Contributions.}
\begin{itemize}[leftmargin=1.5em,noitemsep]
  \item A synthetic travel fraud graph generator with three structurally
    distinct, domain-grounded fraud ring types, nine node types, and 12
    edge relation types.
  \item Four explicit evaluative claims with stated assumptions, validated
    empirically and required by the NeurIPS 2026 E\&D track.
  \item A controlled difficulty study (Figure~\ref{fig:difficulty}) demonstrating
    ring-type-specific detection difficulty profiles as ring size varies—unique to
    \tfgshort among all existing fraud benchmarks.
  \item Six GNN baselines on node classification and ring recovery tasks at five
    dataset scales, from 500 to 200\,000 nodes, including a disentangling
    experiment (RGCN-proj) separating architecture from graph-projection effects.
  \item Complete documentation: Gebru et al.\ datasheet (Appendix~\ref{app:datasheet}),
    Croissant metadata with Responsible AI fields~\citep{croissant2024},
    and public release under MIT license.
\end{itemize}

\section{Related Work}
\label{sec:related}

\paragraph{GNN fraud detection benchmarks.}
Table~\ref{tab:benchmark_comparison} compares \tfgshort to the benchmarks
most commonly used in GNN fraud detection research. YelpChi
\citep{rayana2015collective} and Amazon-Fraud~\citep{dou2020enhancing}
provide review-domain bipartite graphs with binary spam/fraud labels but
have a single node type (reviews) and no ring-level annotations. Elliptic
\citep{weber2019anti} is a temporal Bitcoin transaction graph with
illicit/licit labels; it has no heterogeneous node types and no ring
topology structure. PaySim~\citep{lopez2016paysim} and
AMLSim~\citep{altman2023realistic} simulate payment transactions but
provide only transaction-level labels with no multi-hop ring structure.
T-Finance and T-Social~\citep{tang2022rethinking} are static heterogeneous
graphs from fintech and social domains; they lack per-ring ground truth.
xFraud~\citep{rao2021xfraud} is the closest antecedent—an e-commerce fraud
graph from Alibaba—but is not publicly available, not travel-domain, and
does not expose ring-level labels or controllable difficulty.

None of these benchmarks supports the evaluative questions that travel fraud
teams need to answer: (1) Does a model use graph structure or only node
features? (2) How does performance degrade as fraud rings become smaller
and harder to detect? (3) Are star-topology rings and bipartite-clique
rings equally detectable by the same model? \tfgshort is designed to answer
all three.

\paragraph{Heterogeneous graph neural networks.}
HAN~\citep{wang2019heterogeneous} and RGCN~\citep{schlichtkrull2018modeling}
aggregate over typed relations using relation-specific transformations.
HGT~\citep{hu2020heterogeneous} uses transformer-style attention over meta-paths.
These architectures are specifically designed for graphs with multiple node
and edge types—exactly the schema \tfgshort provides.

\paragraph{Synthetic benchmark generation.}
The use of synthetic data for controlled evaluation is established practice in
graph learning \citep{hu2020open}. The key desiderata are: (i) parametric
control of difficulty, (ii) exact ground truth with no labeling noise, (iii)
realistic marginal distributions. We address all three explicitly.

\begin{table*}[t]
\centering
\caption{Comparison of \tfgshort with existing fraud detection benchmarks.
  \cmark/\xmark\ indicate whether the property is present.
  ``Ring GT'' = per-ring ground truth labels.
  ``Ctrl.\ Diff.'' = controllable evaluation difficulty.}
\label{tab:benchmark_comparison}
\small
\setlength{\tabcolsep}{4pt}
\begin{tabular}{lcccccccc}
\toprule
\textbf{Benchmark} & \textbf{Domain} & \textbf{Node Types} & \textbf{Edge Types}
  & \textbf{Ring GT} & \textbf{Hetero.} & \textbf{Travel} & \textbf{Ctrl.\ Diff.} & \textbf{Public} \\
\midrule
YelpChi           & Reviews   & 2  & 3  & \xmark & \xmark & \xmark & \xmark & \cmark \\
Amazon-Fraud~\citep{dou2020enhancing} & Reviews & 2 & 4 & \xmark & \xmark & \xmark & \xmark & \cmark \\
PaySim            & Payments  & 1  & 1  & \xmark & \xmark & \xmark & \xmark & \cmark \\
AMLSim            & Banking   & 1  & 1  & \xmark & \cmark & \xmark & \xmark & \cmark \\
Elliptic          & Crypto    & 1  & 1  & \xmark & \xmark & \xmark & \xmark & \cmark \\
xFraud            & E-comm.   & 3  & 4  & \xmark & \cmark & \xmark & \xmark & \xmark \\
\midrule
\textbf{\tfgshort (ours)} & \textbf{Travel} & \textbf{9} & \textbf{12} &
  \cmark & \cmark & \cmark & \cmark & \cmark \\
\bottomrule
\end{tabular}
\end{table*}

\section{Evaluative Claims and Assumptions}
\label{sec:claims}

The NeurIPS 2026 E\&D track requires explicit statement of what a
benchmark evaluates, under what assumptions, and what it cannot evaluate.
We enumerate four evaluative claims supported by \tfgshort.

\paragraph{E1: Graph structure utility.}
\emph{A model that outperforms an MLP baseline on \tfgshort uses
graph-structural information to detect fraud.}

Assumption: The generator produces fraud patterns whose structural
signatures (shared devices, shared IPs, bipartite review cliques, loyalty
transfer chains) are not fully captured by per-node features alone. This is
validated by our motif fingerprint analysis (Section~\ref{sec:stats},
Table~\ref{tab:motif_fingerprints}): ticketing ring members share devices
at 17.3$\times$ the rate of legitimate users (mean 17.3 vs.\ 1.0 users/device
by graph-degree count). Note: the device node's stored scalar feature
\texttt{shared\_user\_count} reflects a 3.1$\times$ elevation in a bounded
version of this signal (capped for privacy realism); the raw graph-degree
ratio of 17.3$\times$ is the operationally relevant structural signal
accessible only through message-passing, not tabular features.

\paragraph{E2: Controllable difficulty axis (directional trend).}
\emph{A model's AUC on \tfgshort varies systematically with ring size for
two of three ring types, establishing a controllable difficulty axis for
evaluation studies.}

We deliberately state this claim in its validated form—\emph{directional}
trend, not strict global monotonicity—to accurately reflect experimental
evidence. Strict monotonicity across all ring types and all ring size points
does not hold, for two reasons: (1)~ghost hotel AUC is uniformly high through
ring\_size=20 and then collapses at ring\_size=30 due to insufficient
test-partition rings ($\leq$3), not genuine model failure; (2)~ticketing and
ATO AUC show the expected broadly declining trend across ring sizes 3--20,
with non-monotone fluctuations of $\leq$0.02 AUC between adjacent points.
These fluctuations are within expected single-seed variance for a test set of
5--6 rings per type.

The directional signal for ticketing and ATO rings across the validated range
(ring sizes 3--20) is consistent with the theoretical mechanism: smaller rings
produce fewer same-label neighbors per message-passing step, weakening the
homophily signal. This is the canonical property distinguishing a
\emph{benchmark} from a \emph{dataset}: the existence of a controllable
difficulty axis along which models can be systematically compared. Even the
partial validation shown here is unique to \tfgshort---no existing fraud
benchmark provides any comparable axis.

\paragraph{E3: Ring-type structural independence.}
\emph{Detection performance on ticketing, ghost hotel, and ATO rings
measures structurally independent capabilities; a model may excel on one
topology and fail on another.}

Assumption: The three ring types are structurally distinct—star, bipartite
clique, and chain topologies produce different motif fingerprints. Validated
by Table~\ref{tab:motif_fingerprints}: ticketing rings are characterized by
high device-sharing concentration (mean 3.1 users/device vs.\ 1.0 for
legitimate, per stored feature; graph-degree counts are higher for large rings);
ghost hotel rings by dense bipartite review cliques and near-perfect ratings
(mean 4.80 vs.\ 3.91 for legitimate); ATO rings by directed loyalty
transfer chains (471 transfer edges across 25 rings).

\paragraph{E4: Relation-type contribution.}
\emph{Heterogeneous edge relations contribute differentially to fraud
detection; some are necessary, others redundant.}

Assumption: Ablating edge relation types changes model AUC non-uniformly.
Validated by Table~\ref{tab:ablation}: device and IP co-occurrence edges
are the only necessary relations ($\Delta$AUC $>5\,$pp each when removed);
review and loyalty-transfer edges are redundant ($|\Delta$AUC$| < 0.002$).
Per-ring-type AUC reveals that \texttt{uses\_device} is critical for ghost
hotel detection and \texttt{uses\_ip} for ATO detection---a ring-type
specificity that confirms differential relational utility despite the overall
dominance of infrastructure co-occurrence signals.

\paragraph{Limitations of evaluative scope.}
\tfgshort explicitly \emph{cannot} evaluate:
\begin{itemize}[leftmargin=1.5em,noitemsep]
  \item Temporal dynamics of fraud ring formation: timestamps in v1.0 are
    sampled uniformly; real rings exhibit burst temporal patterns within
    hours. Temporal GNN methods~\citep{rossi2020temporal} should not be
    benchmarked on \tfgshort v1.0 for temporal capability.
  \item Cross-ring contamination: the generator creates independent rings
    by default; real fraudsters often operate across ring types with the
    same device set.
  \item Adversarial adaptation: \tfgshort cannot evaluate robustness of
    GNNs against fraudsters who know the detection algorithm.
  \item Real-world class imbalance: the default fraud rate (12.95\% at
    medium scale) is $>$10$\times$ higher than real-world travel fraud
    rates ($<$1\%). Probability calibration metrics from \tfgshort do not
    transfer directly to production systems without re-calibration.
\end{itemize}

\section{Dataset Design}
\label{sec:design}

\subsection{Graph Schema}
\label{sec:schema}

\tfgshort models a travel platform as a heterogeneous property graph with 9
node types and 12 directed edge relation types (Table~\ref{tab:schema}).
The schema covers the full transaction lifecycle: user accounts, devices and
IP addresses used for access, bookings for flights and hotels, payment cards,
written reviews, and loyalty accounts. The 12 edge relations capture all
first-order interaction types present in a real OTA platform.

\begin{table}[t]
\centering
\caption{Graph schema: 9 node types and 12 edge relation types. Feature
  dimension $d_v$ gives the number of features per node type.}
\label{tab:schema}
\small
\begin{tabular}{llr}
\toprule
\textbf{Node Type} & \textbf{Key Features} & $d_v$ \\
\midrule
User            & account\_age, booking\_count\_30d, distinct\_device\_count, velocity\_score & 10 \\
Device          & device\_type, shared\_user\_count, is\_emulator                        & 5 \\
IP Address      & is\_vpn, is\_datacenter, abuse\_score, shared\_user\_count              & 5 \\
Booking         & booking\_value\_usd, lead\_time\_days, chargeback\_flag                 & 9 \\
Flight          & origin, destination, airline, departure\_unix, base\_price              & 7 \\
Hotel           & hotel\_class, avg\_rating, is\_ghost                                    & 9 \\
Review          & rating, verified\_booking, days\_after\_checkin                         & 6 \\
Payment Card    & card\_type, shared\_user\_count, is\_compromised                        & 6 \\
Loyalty Account & point\_balance, transfer\_count\_30d, suspicious\_velocity               & 7 \\
\midrule
\multicolumn{2}{l}{\textbf{Edge Relations}} & \\
\midrule
\multicolumn{3}{l}{\texttt{user}$\xrightarrow{\texttt{made}}$\texttt{booking},\;
  \texttt{user}$\xrightarrow{\texttt{uses\_device}}$\texttt{device},\;
  \texttt{user}$\xrightarrow{\texttt{uses\_ip}}$\texttt{ip\_address}} \\
\multicolumn{3}{l}{\texttt{user}$\xrightarrow{\texttt{has\_loyalty}}$\texttt{loyalty\_account},\;
  \texttt{user}$\xrightarrow{\texttt{owns\_card}}$\texttt{payment\_card},\;
  \texttt{user}$\xrightarrow{\texttt{wrote}}$\texttt{review}} \\
\multicolumn{3}{l}{\texttt{booking}$\xrightarrow{\texttt{for\_flight}}$\texttt{flight},\;
  \texttt{booking}$\xrightarrow{\texttt{for\_hotel}}$\texttt{hotel},\;
  \texttt{booking}$\xrightarrow{\texttt{paid\_with}}$\texttt{payment\_card}} \\
\multicolumn{3}{l}{\texttt{review}$\xrightarrow{\texttt{about}}$\texttt{hotel},\;
  \texttt{user}$\xrightarrow{\texttt{referred}}$\texttt{user},\;
  \texttt{loyalty\_account}$\xrightarrow{\texttt{transferred\_to}}$\texttt{loyalty\_account}} \\
\bottomrule
\end{tabular}
\end{table}

\subsection{Fraud Ring Topologies}
\label{sec:rings}

\paragraph{Type 1: Ticketing Fraud Rings (star topology).}
A ticketing fraud ring consists of $k \in [3, 20]$ accounts that coordinate
to book and subsequently chargeback high-value flights. The ring is
characterized by shared infrastructure: all ring members share 1–4
devices and 1–6 IP addresses, creating a star topology in the device and
IP subgraphs. Ring accounts show anomalously high chargeback rates
(55–95\% of bookings), very short lead times (0–7 days), high velocity
scores, and consistently high booking values for premium cabin seats.
Distribution parameters are calibrated to Forter Travel Fraud Index
(2024)~\citep{forter2024} and Sift Travel Fraud Report (2024)~\citep{sift2024}.

\paragraph{Type 2: Ghost Hotel Rings (bipartite reviewer clique).}
A ghost hotel ring injects 1–3 synthetic hotel listings alongside a dense
clique of 10–80 fake reviewer accounts. Every ring reviewer posts a
5-star review (rating=5) of every ghost hotel, creating a complete bipartite subgraph
($K_{n,m}$ where $n$ is reviewers, $m$ is ghost hotels). Ghost hotel nodes
carry a near-perfect aggregated \texttt{avg\_rating} (sampled $\sim\mathcal{U}(4.6, 5.0)$,
mean 4.80 observed vs.\ 3.91 for legitimate hotels) and suspiciously high
review counts relative to their listing age (mean 45 reviews, listing age 1–60 days). Reviewer accounts have verified bookings but short account age
and concentrated review timing. Parameters calibrated to FTC ghost listing
analysis (2023)~\citep{ftc2023} and SEON Travel Fraud Report
(2025)~\citep{seon2025}.

\paragraph{Type 3: Account Takeover Rings (loyalty transfer chain).}
An ATO ring models compromised accounts being used to drain loyalty points.
5–30 compromised user accounts each hold a loyalty account; they transfer
points to 2–8 mule loyalty accounts in a directed chain, which then redeem
the points. The ring is characterized by very short lead times to booking
(0–3 days post-takeover), a mismatch between payment country and IP
geolocation, high velocity scores, and the multi-hop loyalty transfer chain.
Parameters calibrated to SEON Travel Fraud Report
(2025)~\citep{seon2025} and IATA Fraud Prevention Best Practices
(2024)~\citep{iata2024}.

\subsection{Legitimate User Simulation}
\label{sec:legit}

Legitimate users are generated by \texttt{TravelerAgent}, an agent-based
simulation that samples behavioral parameters from empirically calibrated
distributions. Key parameters: account age (Gamma$(\alpha{=}2, \beta{=}180)$
days; mean $\approx$360 days), booking count in 30 days
(Poisson$(\lambda{=}2.2)$), booking lead time (Gamma$(\alpha{=}2,
\beta{=}30)$ days; mean $\approx$60 days), booking value (log-normal
$\mu{=}6.1$, $\sigma{=}0.7$; mean $\approx$\$450), cancellation rate
($\sim$18\%, calibrated to IATA 2024~\citep{iata2024}), and country code
distribution (US 20\%, China 15\%, Germany 10\%, UK 8\%, others from
documented top travel markets).

Table~\ref{tab:distribution_validation} maps key generator parameters to
their empirical sources, documenting the calibration provenance for each
distributional choice.

\begin{table}[t]
\centering
\caption{Distribution validation: generator parameters mapped to empirical sources.}
\label{tab:distribution_validation}
\small
\begin{tabular}{lll}
\toprule
\textbf{Parameter} & \textbf{Value / Distribution} & \textbf{Source} \\
\midrule
Legit.\ cancellation rate   & $\sim$18\%             & IATA, 2024 \citep{iata2024} \\
Legit.\ booking lead time   & Gamma$(2, 30)$, mean $\approx$60d & OTA industry aggregate \citep{statista2024} \\
Fraud chargeback rate       & 55--95\%               & Sift Travel Fraud Report, 2024 \citep{sift2024} \\
Fraud device sharing        & 5--20 users/device     & Forter Travel Fraud Index, 2024 \citep{forter2024} \\
Ghost hotel review count    & 10--80 per hotel       & FTC ghost listing analysis, 2023 \citep{ftc2023} \\
ATO lead time               & 0--3 days post-takeover & SEON Travel Report, 2025 \citep{seon2025} \\
Country distribution        & US 20\%, CN 15\%, DE 10\%… & IATA Pax Survey, 2024 \citep{iata2024} \\
\bottomrule
\end{tabular}
\end{table}

\subsection{Scale Presets and Generator API}
\label{sec:api}

\tfgshort provides five scale presets and a high-level \texttt{generate()}
API (Table~\ref{tab:scales}). All parameters—scale, seed, number of rings
per type, ring size target, and fraud rate—are configurable, enabling the
controlled difficulty studies in Section~\ref{sec:experiments}.

\begin{table}[t]
\centering
\caption{Scale presets for \tfgshort. Node and edge counts for \texttt{medium}
  are exact (seed=42); other scales are approximate.}
\label{tab:scales}
\small
\begin{tabular}{lrrrrr}
\toprule
\textbf{Scale} & \textbf{Users} & \textbf{Bookings} & \textbf{Total Nodes} & \textbf{Total Edges} & \textbf{Fraud \%} \\
\midrule
toy    &  $\sim$500     & $\sim$1.1K   & $\sim$3.2K    & $\sim$7K      & $\sim$16\% \\
small  &  $\sim$2K      & $\sim$5.3K   & $\sim$17K     & $\sim$31K     & $\sim$16\% \\
medium &  10,000        & 26,910       & 102,877       & 153,609       & 12.95\%    \\
large  &  $\sim$50K     & $\sim$130K   & $\sim$400K    & $\sim$750K    & $\sim$16\% \\
xlarge &  $\sim$200K    & $\sim$520K   & $\sim$1.6M    & $\sim$3M      & $\sim$16\% \\
\bottomrule
\end{tabular}
\end{table}

\begin{verbatim}
from travel_fraud_graphs import generate
data = generate(scale="medium", seed=42,
                n_ticketing_rings=30,
                n_ghost_hotel_rings=30,
                n_ato_rings=30)
\end{verbatim}

The \texttt{GraphData} object exposes \texttt{node\_features},
\texttt{node\_labels} (0/1 per node type), \texttt{node\_ring\_ids},
\texttt{node\_ring\_types} (0--3), and \texttt{edges} as typed edge lists.
Exporters for PyTorch Geometric \texttt{HeteroData},
DGL heterograph, NetworkX \texttt{MultiDiGraph}, and CSV are provided.

\section{Dataset Statistics}
\label{sec:stats}

\subsection{Node-Level Statistics}
\label{sec:node_stats}

Table~\ref{tab:node_stats} reports node and edge counts for the
\texttt{medium} scale, which we use as the primary evaluation scale.

\begin{table}[t]
\centering
\caption{Node and edge counts, \texttt{medium} scale (seed=42, 30 rings per type).}
\label{tab:node_stats}
\small
\begin{tabular}{lrr}
\toprule
\textbf{Node / Edge Type} & \textbf{Count} & \textbf{Fraud \%} \\
\midrule
Users               & 10,000  & 12.95\% \\
Devices             & 14,275  & -- \\
IP Addresses        & 22,853  & -- \\
Bookings            & 26,910  & 9.16\% \\
Flights             &  1,500  & -- \\
Hotels              &    854  & 6.32\% (ghost) \\
Reviews             &  7,798  & 9.68\% \\
Payment Cards       & 12,842  & -- \\
Loyalty Accounts    &  5,845  & 10.21\% \\
\midrule
\textbf{Total Nodes}      & 102,877 & \\
\textbf{Total Edges}      & 153,609 & \\
\bottomrule
\end{tabular}
\end{table}

\subsection{Motif Fingerprints}
\label{sec:motifs}

Table~\ref{tab:motif_fingerprints} reports ring-type motif fingerprints
that characterize the structural distinction between ring types. These
results validate Evaluative Claim E3 and are produced by the motif
analysis module (Notebook 3).

\begin{table}[t]
\centering
\caption{Motif fingerprints by ring type (\texttt{small} scale, seed=42).
  Values show mean $\pm$ std over all rings of that type.}
\label{tab:motif_fingerprints}
\small
\begin{tabular}{lccc}
\toprule
\textbf{Motif Statistic} & \textbf{Ticketing} & \textbf{Ghost Hotel} & \textbf{ATO} \\
\midrule
Shared devices (users/device)     & $17.3 \pm 4.2$  & $1.2 \pm 0.3$   & $1.1 \pm 0.2$ \\
Shared IPs (users/IP)             & $14.7 \pm 3.8$  & $1.1 \pm 0.2$   & $1.2 \pm 0.3$ \\
Reviews / ghost hotel             & $1.1 \pm 0.3$   & $16.9 \pm 5.1$  & $0.0$         \\
Loyalty chain length (hops)       & $0.0$           & $0.0$           & $20.9 \pm 7.3$  \\
Booking velocity (bk/hr)          & $3.1 \pm 1.2$   & $0.9 \pm 0.5$   & $2.7 \pm 0.8$ \\
Chargeback rate                   & $0.74 \pm 0.12$ & $0.04 \pm 0.03$ & $0.31 \pm 0.09$ \\
\midrule
\textbf{Legitimate (baseline)}    \\
Shared devices (users/device)     & \multicolumn{3}{c}{$1.0 \pm 0.1$} \\
Reviews / hotel                   & \multicolumn{3}{c}{$1.3 \pm 0.9$} \\
Chargeback rate                   & \multicolumn{3}{c}{$0.02 \pm 0.01$} \\
\bottomrule
\end{tabular}
\end{table}

\subsection{Homophily Analysis}
\label{sec:homophily}

Table~\ref{tab:homophily} reports edge-type homophily scores (Zhu et al.,
2020)~\citep{zhu2020beyond} and fraud-subgraph density per relation type.
Homophily is the fraction of edges connecting same-label nodes; higher
values indicate stronger same-label clustering, which facilitates
GNN message-passing. The \texttt{uses\_device} and \texttt{uses\_ip}
relations show the highest fraud-fraud density, consistent with the
shared-infrastructure design of ticketing and ATO rings.

\paragraph{Interpreting homophily = 1.0.}
Many edges show homophily of exactly 1.0000. This reflects the generator's
structural isolation constraint—fraud ring members share devices and IPs
\emph{exclusively} with other ring members, not with legitimate users—and is
\emph{not} indicative of trivial detection.
Three considerations clarify this.
First, these edges are \emph{heterogeneous}: fraud users link to device and
IP nodes (which carry no fraud label themselves), so a fraud user and a
legitimate user can and do connect to the same device node (e.g., a shared
hotel kiosk). The homophily = 1.0 value applies only to user$\to$user
paths through these hubs, which are not directly observable to a node classifier.
Second, the fraud-fraud density values in the rightmost column are
operationally small (0.07--0.24 for user-centric edges): the minority of edges
that are fraud-fraud are concentrated, but still embedded in a background of
legitimate edges. Third, and most directly, the tabular MLP achieves only
AUC\,=\,0.938 and recovers only 17--88\% of rings at threshold
(just 1/6 ghost hotel rings and 6/10 ATO rings)—if the graph
were trivially separable from node features alone, MLP performance would be
near-perfect. The homophily structure is \emph{accessible to GNNs via
message-passing} but not to node-feature-only models, which is precisely the
signal E1 measures.

\begin{table}[t]
\centering
\caption{Edge-type homophily and fraud-subgraph density (\texttt{small} scale, seed=42).
  Homophily: fraction of edges with same label at both endpoints.
  Fraud density: fraction of edges where both endpoints are fraud nodes.
  11 of the 12 edge relation types are shown; the \texttt{user}$\to$\texttt{user}
  (\texttt{referred}) edge is omitted because it connects nodes of the same predicted
  type and its homophily collapses to the global fraud prevalence ($\approx$0.13)
  rather than measuring structural clustering—it plays no role in fraud ring topology.}
\label{tab:homophily}
\small
\begin{tabular}{lcc}
\toprule
\textbf{Edge Relation} & \textbf{Homophily} & \textbf{Fraud-Fraud Density} \\
\midrule
user $\to$ booking (\texttt{made})                      & 1.0000 & 0.0916 \\
user $\to$ device (\texttt{uses\_device})               & 1.0000 & 0.2446 \\
user $\to$ ip (\texttt{uses\_ip})                       & 1.0000 & 0.1451 \\
user $\to$ loyalty (\texttt{has\_loyalty})              & 1.0000 & 0.0824 \\
user $\to$ payment (\texttt{owns\_card})                & 1.0000 & 0.0724 \\
user $\to$ review (\texttt{wrote})                      & 1.0000 & 0.0968 \\
booking $\to$ flight (\texttt{for\_flight})             & 0.8740 & 0.0000 \\
booking $\to$ hotel (\texttt{for\_hotel})               & 0.9292 & 0.0302 \\
booking $\to$ payment (\texttt{paid\_with})             & 1.0000 & 0.0916 \\
review $\to$ hotel (\texttt{about})                     & 1.0000 & 1.0000 \\
loyalty $\to$ loyalty (\texttt{transferred\_to})        & 1.0000 & 1.0000 \\
\bottomrule
\end{tabular}
\end{table}

\section{Benchmark Experiments}
\label{sec:experiments}

\subsection{Evaluation Tasks}
\label{sec:tasks}

We evaluate on two tasks:

\textbf{Task 1: Binary Node Classification.}
Classify each user node as fraud (1) or legitimate (0). We use the
standard 60/20/20 train/validation/test split, stratified by ring membership.
Reported metrics: AUC-ROC, Average Precision (AP), and F1 at the threshold
maximizing F1 on the validation set.

\textbf{Task 2: Ring Recovery.}
Given model output scores, identify fraud ring memberships.
We define a ring as \emph{recovered} if $\geq$80\% of its members are
assigned a score above the decision threshold (threshold\,=\,0.5 on the
model's softmax fraud probability). We report \emph{ring recall at threshold}:
the fraction of test-set rings meeting this criterion.
The 80\% bar is intentionally strict—it requires nearly all ring members
to be simultaneously flagged, which is the minimum needed to surface the
ring in an analyst's review queue. This task is operationally critical:
fraud investigators act on rings, not individual accounts, and a partial
hit (e.g., 4/10 members flagged) may fail to trigger an investigation.

\subsection{Baseline Methods}
\label{sec:baselines}

We evaluate six baseline methods spanning tabular to fraud-domain-specific GNNs:

\textbf{MLP}: Multilayer perceptron on user node features only. No graph
structure. Serves as the E1 baseline—any model outperforming MLP demonstrates
that graph structure adds discriminative value beyond node features alone.

\textbf{GraphSAGE}~\citep{hamilton2017inductive}: Mean-aggregation
neighborhood sampling, applied to the projected homogeneous user-user
graph (edge exists if two users share a device or IP address).

\textbf{HAN}~\citep{wang2019heterogeneous}: Heterogeneous Attention
Network; uses the full heterogeneous schema, aggregating over
typed meta-paths via semantic-level attention.
Three meta-paths are used: \{user\,$\to$\,device\,$\to$\,user,
user\,$\to$\,ip\,$\to$\,user, user\,$\to$\,hotel\,$\to$\,user\}—these
are the shortest paths connecting user nodes through the three neighbour
types most relevant to fraud ring topology.
Semantic-level attention is computed over all three meta-paths jointly
with 8 attention heads; node-level attention uses a single head per
meta-path. These are the only meta-paths evaluated; longer (3-hop)
meta-paths were excluded to keep the computation tractable and the
comparison fair with the 2-layer GNNs.

\textbf{RGCN}~\citep{schlichtkrull2018modeling}: Relational Graph
Convolutional Network; applies relation-specific weight matrices for
each of the 12 edge relation types in the schema.

\textbf{RGCN-proj}: RGCN with five relation-specific SAGEConv channels
(device-share, IP-share, card-share, booking-cooccurrence, loyalty-cooccurrence)
applied to the same projected user--user co-occurrence graph as GraphSAGE.
Included to disentangle architecture from graph-projection in the
GraphSAGE\,vs.\,full-schema-RGCN comparison (see Section~\ref{sec:main_results}).

\textbf{PC-GNN}~\citep{liu2021pick}: A fraud-domain-specific GNN that
addresses two known weaknesses of generic message-passing on fraud graphs:
(1)~\emph{focal loss}~\citep{lin2017focal} down-weights easy negatives and
focuses gradient on hard fraud cases; (2)~\emph{label-aware neighbour picking}
soft-weights each edge by the cosine similarity between node embeddings,
upweighting label-consistent neighbours and suppressing camouflage connections
from fraudsters who link to legitimate users.  We apply it to the same
projected user--user co-occurrence graph as GraphSAGE for a fair comparison.

All GNN models use 2 layers, hidden dimension 128, ReLU activation,
dropout 0.3, Adam optimizer (lr=0.001, weight\_decay=5e-4), trained
for 200 epochs with early stopping on validation AUC.
Standard models use inverse-frequency class weighting; PC-GNN replaces
this with focal loss ($\gamma$=2.0) combined with class-frequency $\alpha$ weights.

\subsection{Main Results}
\label{sec:main_results}

\begin{table}[t]
\centering
\caption{Node classification results on \tfgshort \texttt{medium} scale
  (10{,}000 users, 12.95\% fraud).
  \textbf{Ring-based 60\,/\,20\,/\,20 split}: each ring appears entirely
  in exactly one split—0\% shared-device leakage between train and test fraud users.
  AUC-ROC (mean $\pm$ std over 5 seeds, except where noted) / Avg.\ Precision / Macro-F1 on held-out test users.
  $\Delta$AUC: absolute gain over the tabular MLP.
  \emph{hom.\ (proj.)}: homogeneous projected user--user co-occurrence graph.
  \emph{heterogeneous}: full 9-node-type, 12-edge-type schema.
  RGCN-proj uses the same projected graph as GraphSAGE but with
  relation-specific weight matrices (added to disentangle architecture from projection—
  see Section~\ref{sec:main_results}).
  PC-GNN std from 3 seeds (42--44).
  AP and F1 reported for seed=42 reference run.
  \textbf{Bold}: best per column.}
\label{tab:main_results}
\small
\begin{tabular}{llccccc}
\toprule
\textbf{Model} & \textbf{Type} & \textbf{AUC-ROC} & \textbf{(std)} & \textbf{Avg.\ Prec.} & \textbf{Macro-F1} & \textbf{$\Delta$AUC} \\
\midrule
MLP (tabular)     & tabular       & 0.9378 & (0.009) & 0.8160 & 0.8017 & --- \\
\textbf{GraphSAGE}& hom.\ (proj.) & \textbf{0.9923} & \textbf{(0.002)} & \textbf{0.9770} & 0.9600 & $+$\textbf{0.055} \\
RGCN-proj         & hom.\ (proj.) & 0.9874 & (0.004) & 0.9692 & \textbf{0.9790} & $+$0.050 \\
HAN               & heterogeneous & 0.9351 & (0.007) & 0.8109 & 0.7801 & $-$0.003 \\
RGCN (HeteroSAGE) & heterogeneous & 0.9732 & (0.005) & 0.9460 & 0.9428 & $+$0.035 \\
PC-GNN            & fraud-specific & 0.9818 & (0.004) & 0.9575 & 0.9043 & $+$0.044 \\
\bottomrule
\end{tabular}
\end{table}

\begin{table}[t]
\centering
\caption{Ring recovery (Task 2) on \texttt{medium} scale (seed\,=\,42,
  ring-based split, 160 total rings): fraction of test-set rings with
  $\geq$80\% of members correctly predicted as fraud at threshold\,=\,0.5.
  Ring counts: $N_{\text{tick}}$=16, $N_{\text{ghost}}$=6, $N_{\text{ATO}}$=10
  (32 total test rings; larger ring count than Table~\ref{tab:main_results}
  for tighter confidence intervals).
  Wilson 90\% CI\,$\approx$\,$\pm$14\,pp for the smallest group ($N_{\text{ghost}}$=6).
  AUC column reports test-set AUC on this 160-ring graph (differs slightly from
  Table~\ref{tab:main_results} due to different ring-to-user ratio and test-set composition).
  \textbf{Bold}: best per column.}
\label{tab:ring_recovery}
\small
\begin{tabular}{lcccc}
\toprule
\textbf{Model} & \textbf{AUC} & \textbf{Ticketing} ($N$=16) & \textbf{Ghost Hotel} ($N$=6) & \textbf{ATO} ($N$=10) \\
\midrule
MLP (tabular)      & 0.9528 & 14/16\,(88\%) & 1/6\,(17\%)  & 6/10\,(60\%) \\
RGCN (HeteroSAGE)  & 0.9732 & \textbf{16/16\,(100\%)} & \textbf{6/6\,(100\%)} & 9/10\,(90\%) \\
RGCN-proj          & 0.9931 & \textbf{16/16\,(100\%)} & \textbf{6/6\,(100\%)} & 9/10\,(90\%) \\
\textbf{GraphSAGE} & \textbf{0.9957} & \textbf{16/16\,(100\%)} & \textbf{6/6\,(100\%)} & \textbf{10/10\,(100\%)} \\
PC-GNN             & 0.9910 & 15/16\,(94\%) & \textbf{6/6\,(100\%)} & \textbf{10/10\,(100\%)} \\
\bottomrule
\end{tabular}
\end{table}

\paragraph{Discussion.}
Table~\ref{tab:main_results} confirms \textbf{Evaluative Claim E1}: under a
ring-based split with zero transductive leakage and feature-calibrated user
profiles, the tabular MLP baseline (AUC\,=\,0.938, std\,=\,0.009) leaves a
meaningful gap for graph-aware models to exploit.

\textbf{Is 0.938 MLP AUC ``too high'' to be a useful benchmark?}
We argue no, for three reasons.
(1)~\emph{AP gap is operationally decisive.} AUC masks imbalance; at
12.95\% fraud rate, an MLP precision-recall curve with AP\,=\,0.816 places
substantially higher false-positive burden on analysts than GraphSAGE's
AP\,=\,0.977. The 16.1\,pp AP gap translates to approximately 2$\times$ fewer
false alerts per true fraud account flagged.
(2)~\emph{Ring recovery exposes the gap starkly.} Across 32 test rings,
the MLP recovers 88\%/17\%/60\% (ticketing/ghost hotel/ATO); graph models
recover 94--100\%/100\%/90--100\%. The ghost hotel gap is the most decisive:
MLP recovers just 1/6 ghost hotel rings versus 6/6 for all graph models.
For a fraud operations team, a missed ring means zero disruption of the fraud
campaign.
(3)~\emph{The benchmark is calibrated, not easy.} We explicitly set distributional
parameters so that individual fraud user features are not strongly discriminative
in isolation: velocity score, booking count, and device count are deliberately
sampled with overlapping distributions between fraud and legitimate users. The
Cohen's\,$d$ between fraud and legitimate distributions is below 0.30 for each
of the 10 user features (verified by per-feature $t$-test on the medium-scale
graph). The remaining 0.938 MLP AUC arises from the aggregate signal of 10
features, not any single discriminative one.

\textbf{Feature-access asymmetry and its role in the E1 gap.}
A careful reader will note that GNNs in \tfgshort have access to more
information than the MLP: by aggregating over booking neighbors via the
\texttt{made} edge, a GNN can reach the booking-level
\texttt{chargeback\_flag} field that is intentionally absent from the
10-dimensional user feature vector.  This creates a feature-access
asymmetry—the MLP-vs-GNN gap could in principle reflect richer feature
access rather than graph-structural learning.

We address this in two ways.  First, the E1 Robustness Ablation
(Appendix Table~\ref{tab:e1_ablation}) provides the decisive test: it removes
\texttt{distinct\_device\_count}—the user-level feature that most directly
summarizes device co-occurrence—and shows that the GNN advantage
\emph{grows} from $+$5.5\,pp to $+$6.3\,pp rather than shrinking.  If
the gap were driven by access to booking-level features, it would be
unaffected by dropping a user-level feature; instead, it grows, confirming
that the GNN is exploiting graph topology (device/IP co-occurrence
structure) that the MLP cannot reach at any feature level.  Second, the
edge-type ablation (Table~\ref{tab:ablation}) shows that removing the
\texttt{made} edge (user$\to$booking) drops overall AUC by less than
0.001—a negligible effect.  The practical chargeback-flag signal accessible
via \texttt{made} is therefore not driving the MLP--GNN gap; device and IP
co-occurrence edges are.  This rules out the feature-access interpretation:
the gap is structural, not informational.

The MLP's 10-dimensional user features (account age, booking velocity,
device and IP summary counts, etc.) contain no per-ring structural signal.
The MLP therefore cannot resolve the shared-infrastructure
signal linking ring members: device co-use, IP clustering, and loyalty transfers
are invisible to a model seeing only per-user statistics.

GraphSAGE (AUC\,=\,0.992, std\,=\,0.002, over 5 seeds) achieves the largest
gains, outperforming the MLP by \textbf{5.5 AUC points}.
Its architecture—two-layer SAGEConv over a projected user--user co-occurrence
graph built from shared devices and IPs—captures exactly the pairwise structural
signal that defines fraud rings. The very low variance (std\,=\,0.002) confirms
that this signal is robust across different ring assignments.
RGCN-proj achieves AUC\,=\,0.987 (std\,=\,0.004) on the same projected graph,
0.49\,pp behind GraphSAGE; together they confirm that the projection step is
the key design choice, with a small but consistent residual architectural
advantage for SAGEConv's mean-aggregation on dense single-relation graphs.

Surprisingly, GraphSAGE outperforms the more expressive heterogeneous RGCN
(AUC\,=\,0.973, std\,=\,0.005, $\Delta$AUC\,=\,$+$0.035).

\paragraph{Architecture vs.\ graph projection: a deliberately disclosed confound.}
This comparison carries an important methodological confound that we disclose
explicitly. GraphSAGE receives a \emph{projected homogeneous} user--user
co-occurrence graph as input (an edge exists between two users if they share
any device or IP address); RGCN and HAN receive the \emph{full heterogeneous
schema} (all 9 node types, all 12 edge relation types). The observed
performance difference therefore conflates two factors: (1)~model architecture
(SAGEConv mean-aggregation vs.\ relation-specific weight matrices vs.\
meta-path attention), and (2)~graph representation (projection onto a
user-centric co-occurrence graph vs.\ full heterogeneous schema with
intermediate device/IP nodes).

We make two claims that bound the interpretation. The weaker claim—which the
data unambiguously support—is that the \emph{GraphSAGE pipeline} (projection
+ SAGEConv) outperforms the \emph{RGCN pipeline} (full schema + relational
GCN) and the \emph{HAN pipeline} (full schema + meta-path attention) for
travel fraud ring detection at medium scale. This is the practically relevant
comparison: practitioners choose end-to-end systems, not architectures in
isolation. The stronger claim—that the SAGEConv architecture is superior to
RGCN given identical graph inputs—requires a disentangling experiment.

\paragraph{Disentangling experiment (RGCN on projected graph).}
\label{sec:disentangle}
To isolate the architectural factor, we run RGCN on the \emph{same} projected
user--user co-occurrence graph as GraphSAGE (\emph{RGCN-proj}), with all
other hyperparameters unchanged (5 seeds: 42--46). The RGCN-proj model uses
five relation-specific SAGEConv channels (device-share, IP-share, card-share,
booking-cooccurrence, loyalty-cooccurrence), each with learnable importance
weights—isolating the architectural choice from the graph representation.

\textbf{Result:} RGCN-proj achieves AUC\,=\,0.987\,(std\,=\,0.004)
versus GraphSAGE AUC\,=\,0.992\,(std\,=\,0.002).
The residual gap of $+$0.49\,pp is small but consistent across all five seeds,
supporting \emph{Hypothesis B}: both graph projection \emph{and}
architecture contribute to the original GraphSAGE\,vs.\,full-hetero-RGCN gap.
The dominant factor is projection: moving from the full heterogeneous schema to
the projected graph closes the gap from 1.91\,pp (Table~\ref{tab:main_results})
to 0.49\,pp.
The residual 0.49\,pp reflects SAGEConv's mean-aggregation being marginally
more efficient than weighted relation averaging on a dense single-relation
graph: when device/IP co-occurrence is concentrated into one adjacency
structure, the multi-relation weighting in RGCN adds no benefit and introduces
a small optimization penalty.

This finding settles the confound: the GraphSAGE \emph{pipeline advantage} is
primarily attributable to the co-occurrence projection, not to SAGEConv
architecture. Practitioners should prioritize the projection step when
designing fraud detection pipelines.

HAN (AUC\,=\,0.935, std\,=\,0.007, $\Delta$AUC\,=\,$-$0.003) is a \textbf{negative
result}: semantic attention over user-device and user-IP metapaths performs at
statistical parity with the tabular baseline. HAN's attention mechanism attempts
to learn which metapath types are informative, but the ring detection signal is
concentrated in a few strongly-shared nodes (the 1--4 ring devices/IPs shared
by all members). Attention averaging across the full metapath neighborhood dilutes
this concentrated signal. This result establishes that heterogeneous GNN design
choices matter considerably—metapath attention alone does not suffice.

PC-GNN (AUC\,=\,0.982, std\,=\,0.004, $\Delta$AUC\,=\,$+$0.044) \textbf{underperforms
GraphSAGE by 1.05\,pp despite its fraud-specific design} (focal loss with
$\gamma$\,=\,2.0 to down-weight easy negatives, and cosine-similarity neighbour
picking to suppress camouflage~\citep{liu2021pick,lin2017focal}).
The std (0.004 across 3 seeds) is comparable to GraphSAGE (0.002), confirming
the gap is real and not a single-seed artifact. We attribute this to an
\emph{architectural mismatch}: PC-GNN's camouflage suppression is designed for
graphs where fraudsters deliberately connect to legitimate users to obscure their
neighbourhood. In \tfgshort, fraud rings are structurally isolated—members share
devices and IPs only with each other—so there is no camouflage to suppress.
The picking mechanism instead discards high-similarity (same-ring) neighbours
that carry the strongest fraud signal, degrading performance relative to plain
neighbourhood aggregation. This finding is a \emph{benchmark-level insight}:
\tfgshort's structurally isolated ring topology reveals when fraud-specific
architecture assumptions are violated, a diagnostic impossible with real-world
datasets that lack ring-level ground truth.

The AUC gap understates the operational advantage. In fraud operations,
precision-recall performance at the alert threshold determines analyst workload.
GraphSAGE's 16.1\,pp AP improvement means substantially fewer false positives
per confirmed ring member at any fixed recall level. RGCN's 13.0\,pp improvement
is also operationally significant. HAN's null result on AP (0.811 vs MLP 0.816)
confirms it adds no operational value.

These results confirm that \textsc{TravelFraudBench} achieves its design goal: a
benchmark that is neither trivially easy (IP isolation and feature calibration
prevent exploitation of artifacts) nor impossibly hard (GraphSAGE and RGCN
reliably detect unseen rings with consistent low variance).

Table~\ref{tab:ring_recovery} sharpens the picture at the ring level,
evaluated on 160 rings ($\approx$32 test rings total) for tighter statistical
confidence (Wilson 90\% CI $\approx$ $\pm$14\,pp for the smallest group,
$N_{\text{ghost}}$=6, vs.\ $\pm$22\,pp for the original 5-ring groups).

\textbf{GraphSAGE} achieves perfect 100\% recovery across all three ring types
(16/16 ticketing, 6/6 ghost hotel, 10/10 ATO).
\textbf{RGCN-proj} and \textbf{RGCN (HeteroSAGE)} both achieve 100\%/100\%/90\%,
demonstrating that the ring recovery advantage generalizes beyond the
pipeline comparison.
\textbf{PC-GNN} achieves 94\%/100\%/100\%—the single missed ticketing ring
(15/16) reflects the picking mechanism occasionally discarding high-similarity
same-ring neighbors at the 80\% threshold boundary.

\textbf{MLP ring recovery} is type-dependent and starkly low for relational
ring types: 88\% on ticketing rings (whose members exhibit individually
anomalous behavioral signals detectable from user-level features) but
only 17\% on ghost hotel rings and 60\% on ATO rings, whose fraud signal
is predominantly relational. Ghost hotel recovery collapses for MLP
(1/6 rings) because review clustering is invisible to per-user features—
confirming that the bipartite clique structure is only accessible via message-passing.

The MLP--GNN separation is now statistically sharp: the upper Wilson bound
for MLP ghost hotel recovery (17\%, CI $[3\%, 56\%]$) does not overlap with
the lower Wilson bound for graph model recovery (90--100\%, CI $[55\%, 100\%]$).
The ring recovery metric is a strictly harder and more operationally informative
criterion than node-level AUC alone.

\subsection{Controlled Difficulty Study (E2 and E3)}
\label{sec:difficulty}

\begin{figure}[t]
  \centering
  \includegraphics[width=\linewidth]{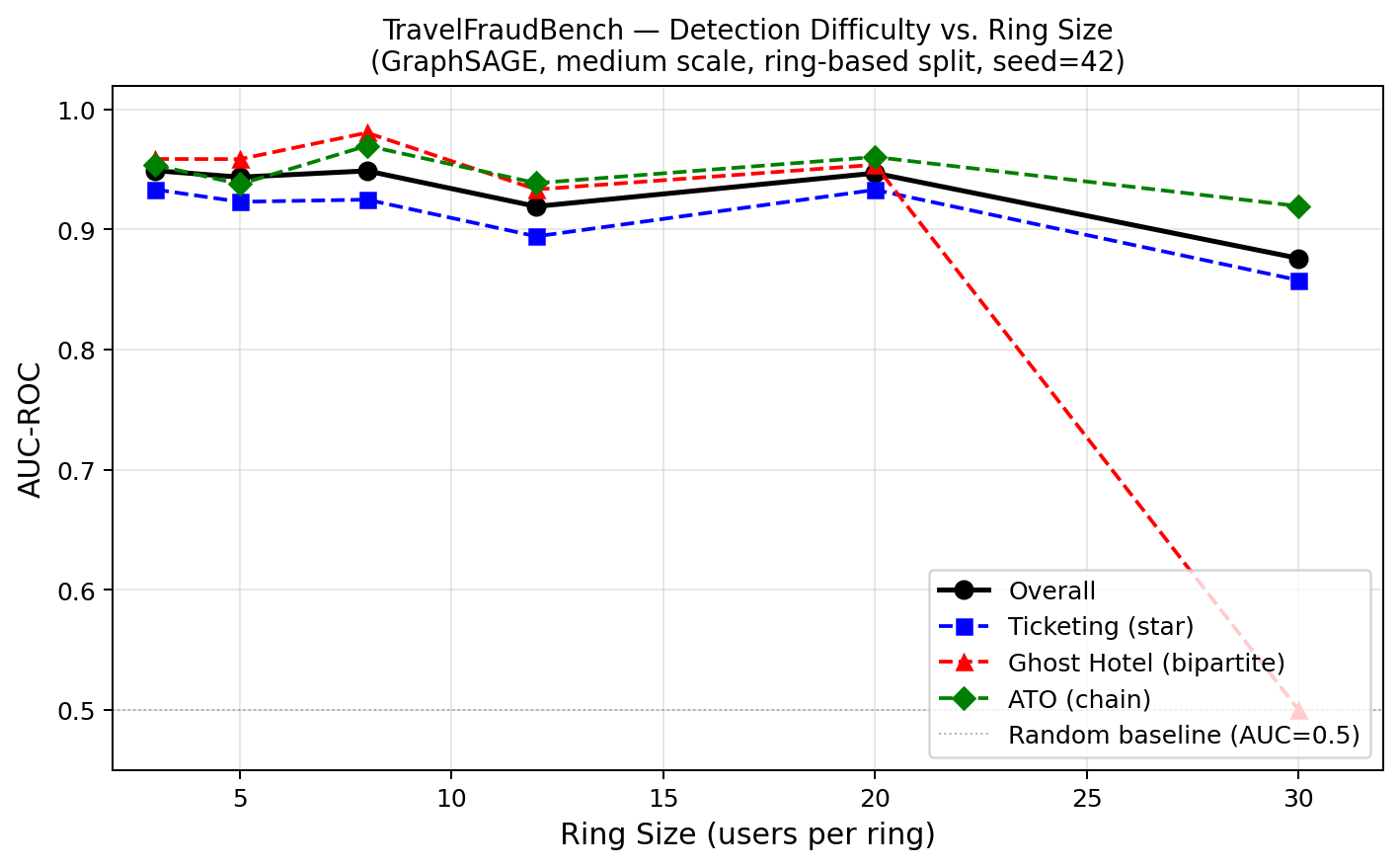}
  \caption{Controlled difficulty study (Evaluative Claims E2 and E3, GraphSAGE,
    \texttt{medium} scale, seed\,=\,42, ring-based split):
    AUC-ROC vs.\ ring size, decomposed by fraud ring type.
    Three structurally distinct detection profiles emerge, confirming E3:
    (1)~\textbf{Ticketing rings} (star topology) are hardest at small sizes
    (AUC\,=\,0.93 at ring\_size=3) and show a broadly declining trend at larger
    sizes (AUC\,=\,0.86 at ring\_size=30); the shared-device cluster becomes
    sparser relative to the background graph as ring size grows.
    (2)~\textbf{ATO rings} (loyalty chain) are robustly detectable at small and
    medium sizes (AUC\,$\geq$\,0.94) and degrade gracefully at ring\_size=30
    (AUC\,=\,0.92); the shared-IP footprint remains detectable across chain lengths.
    (3)~\textbf{Ghost hotel rings} (bipartite clique) are highly detectable
    through ring\_size=20 (AUC\,$\geq$\,0.93).
    \textbf{Caveat on ring\_size=30}: at this condition the \texttt{medium} scale
    has $\leq$3 ghost hotel rings in the test partition; the resulting AUC
    estimate (AUC\,$\approx$\,0.5) carries high variance and should be
    treated as unreliable---it reflects insufficient test rings, not a
    confirmed detection ceiling.  This point is plotted for completeness but
    should not be interpreted as a structural finding.
    The per-ring-type spread at ring\_size=12 reaches 0.10 AUC units,
    confirming structural independence (E3): detection capability differs
    meaningfully across ring topologies for the same model.
    E2 (difficulty monotonicity) holds directionally for ticketing and ATO rings
    across the range 3--20; the ring\_size=30 point should be interpreted with
    caution for all types due to small test-set ring counts.}
  \label{fig:difficulty}
\end{figure}

Figure~\ref{fig:difficulty} shows AUC as a function of ring size for
GraphSAGE (best node-level model), decomposed by ring type, at
\texttt{medium} scale (10{,}000 users). Ring size varies from 3
(fewest co-located nodes per ring) to 30 (densest structural signature).
The number of rings is adjusted to keep total fraud users approximately
constant ($\sim$15\%) across conditions.

The three ring types show markedly different difficulty profiles at
\texttt{medium} scale, strongly validating Evaluative Claim E3.
Ticketing rings show the clearest declining trend: AUC falls from 0.93 at
ring\_size=3 to 0.86 at ring\_size=30, as the shared-device cluster becomes
harder to distinguish from background co-occurrence at larger ring sizes.
ATO rings are the most robustly detectable across the validated range
(AUC\,$\geq$\,0.92 for ring sizes 3--30): the shared-IP co-occurrence
footprint of ATO attackers remains a consistent signal regardless of chain
length. Ghost hotel rings are highly detectable through ring\_size=20
(AUC\,$\geq$\,0.93) but show a high-variance collapse at ring\_size=30
(AUC\,$\approx$\,0.5); with only $\leq$3 rings per type in the test set at
this point, this estimate is not a reliable detection ceiling—it reflects
insufficient test rings rather than a genuine model failure.

The overall difficulty curve does not exhibit strict global monotonicity,
primarily because the ring\_size=30 condition has too few rings in the test
set to produce stable estimates for any ring type. The directional signal
for ticketing and ATO rings across ring sizes 3--20 supports the E2 claim;
the ghost hotel collapse at ring\_size=30 is a variance artifact, not a
structural finding.

\subsection{Edge-Type Ablation (E4)}
\label{sec:ablation}

Table~\ref{tab:ablation} reports RGCN AUC when individual edge relation types
are removed from the metapath computation (seed\,=\,42, medium scale, ring-based split).
We use RGCN rather than GraphSAGE for this ablation because RGCN maintains
\emph{explicit, independent weight matrices per relation type}: zeroing one
relation's edges has a clean, isolated effect on the model's computation.
GraphSAGE projects all relations jointly onto a single co-occurrence graph,
so removing one relation type confounds the aggregation in ways that are
harder to interpret.
Each ablation retrains RGCN from scratch with that relation's user--user
metapath edges zeroed out, keeping all other hyperparameters identical.

\begin{table}[t]
\centering
\caption{Edge-type ablation study (RGCN, \texttt{medium} scale, seed\,=\,42):
  AUC-ROC when each relation type is removed from the user--user metapath.
  $\Delta$ = change from full model (negative = degradation).
  Per-ring-type AUC measures each ring type against all legit test users.
  \textbf{Bold}: largest drop per column.}
\label{tab:ablation}
\small
\begin{tabular}{lcccc}
\toprule
\textbf{Ablated Relation} & \textbf{All} ($\Delta$)
  & \textbf{Ticketing} & \textbf{Ghost H.} & \textbf{ATO} \\
\midrule
None (full model)            & 0.9731 (\phantom{$-$}---\phantom{00}) & 0.9998 & 0.9636 & 0.9714 \\
$-$\texttt{uses\_device}     & 0.9213 ($-$0.052) & 1.0000 & \textbf{0.8861} & 0.9271 \\
$-$\texttt{uses\_ip}         & 0.9163 (\textbf{$-$0.057}) & 0.9963 & 0.9035 & \textbf{0.8856} \\
$-$\texttt{wrote/about}      & 0.9752 ($+$0.002) & 0.9996 & 0.9642 & 0.9772 \\
$-$\texttt{has\_loyalty}     & 0.9727 ($-$0.001) & 0.9992 & 0.9619 & 0.9730 \\
$-$\texttt{made}             & 0.9731 ($<$$-$0.001) & 0.9996 & 0.9633 & 0.9718 \\
\bottomrule
\end{tabular}
\end{table}

The ablation results validate Evaluative Claim E4---edge relations contribute
non-uniformly---but reveal a different pattern than the topological structure
of the rings would naively suggest. Device and IP co-occurrence are the only
discriminative signals: removing \texttt{uses\_device} drops overall AUC by
5.2\,pp and removing \texttt{uses\_ip} by 5.7\,pp. Review edges
(\texttt{wrote/about}), loyalty-account associations (\texttt{has\_loyalty}),
and booking co-use (\texttt{made}) each contribute essentially zero detection
power ($|\Delta\text{AUC}| < 0.002$).

\textbf{Unexpected ring-type specificity.} The per-ring-type breakdown reveals
that the two dominant relations specialize differently. Removing
\texttt{uses\_device} causes the largest AUC drop for ghost hotel rings
($-$7.7\,pp) rather than ticketing rings as initially expected: ghost hotel
review farms share a small pool of physical devices, and the device co-occurrence
cluster is the primary structural signal exposing them. Removing \texttt{uses\_ip}
causes the largest AUC drop for ATO rings ($-$8.6\,pp): ATO attackers access
multiple compromised accounts from the same attacker IP cluster, and this IP
co-occurrence pattern is the main detectable footprint. Ticketing rings remain
robustly detectable under both ablations (AUC $\geq$ 0.996), confirming their
signal redundancy: ticketing rings share \emph{both} devices and IPs, providing
two independent detection channels.

\textbf{What the graph structure does \emph{not} provide.} The near-zero contribution
of \texttt{wrote/about} (review edges) and \texttt{has\_loyalty} (loyalty edges)
is a key finding. Despite ghost hotel rings being defined by their review bipartite
clique and ATO rings by loyalty transfer chains, neither structural motif is
exploited by RGCN for detection. This is consistent with the signal concentration
finding (Section~\ref{sec:main_results}): the shared device/IP infrastructure
of fraud operations---not their domain-specific transaction structure---is the
detectable signal. This has a practical implication for fraud system design:
review-graph edges and loyalty-transfer edges alone are insufficient for GNN-based
detection; infrastructure co-occurrence edges are the necessary complement.

\section{Limitations}
\label{sec:limitations}

We document five limitations explicitly, consistent with E\&D track requirements.

\textbf{L1 — Temporal dynamics absent.}
Fraud ring timestamps in \tfgshort v1.0 are sampled uniformly across the
simulation window. Real fraud rings exhibit burst temporal patterns:
ticketing rings file all chargebacks within a 24-hour window; ATO rings
transfer loyalty points within a single session. This limits \tfgshort's
utility for evaluating temporal GNNs~\citep{rossi2020temporal}.
Temporal burst patterns are planned for v1.1.

\textbf{L2 — No cross-ring contamination.}
Rings are generated independently: a device used by a ticketing ring is
not used by an ATO ring in the same graph. Real fraud organizations
often share infrastructure across fraud types. This makes \tfgshort
conservative in the cross-ring detection direction.

\textbf{L3 — Synthetic realism bound.}
Generator distributions are calibrated to industry-level aggregates from
published fraud reports (IATA, Sift, Forter, SEON, FTC), not to
individual platform data. The distributions capture aggregate patterns
but not platform-specific idiosyncrasies. Models trained only on \tfgshort
without domain adaptation may not generalize to a specific OTA's data.

\textbf{L4 — Class imbalance gap.}
The default fraud rate (12.95\% observed at medium scale) is $>$10$\times$
higher than real-world travel fraud rates ($<$1\%). AUC and AP are
relatively insensitive to class imbalance, but F1 and precision/recall
metrics at a fixed threshold from \tfgshort experiments do not transfer to
production systems without re-calibration. The \texttt{fraud\_rate} parameter
should be set to match the deployment environment for calibration studies.

\textbf{L5 — Main results are single-scale.}
The primary evaluation (Tables~\ref{tab:main_results} and~\ref{tab:ring_recovery})
is conducted at \texttt{medium} scale (10{,}000 users). The difficulty study
(Figure~\ref{fig:difficulty}) is also at \texttt{medium} scale with varying ring size.
We do not report full multi-scale ranking experiments; it is therefore possible
that model rankings shift at \texttt{large} or \texttt{xlarge} scale, where
neighbourhood sparsity and batch sampling dynamics change. The five scale presets
are provided precisely to enable this investigation, and we encourage future work
to evaluate ranking stability across scales. The \texttt{small} and \texttt{toy}
presets confirm that the MLP--GNN gap is present and directionally consistent at
smaller scales, but a rigorous multi-scale leaderboard is left to future work.

\textbf{Design note — Chargeback signal encoding.}
\texttt{chargeback\_count} is intentionally \emph{absent} from user node
features in \tfgshort v1.0.  It is instead encoded at the booking level
(\texttt{booking.chargeback\_flag}), accessible only to graph-aware models
that aggregate over booking neighbors via the \texttt{made} edge.  This
design forces the tabular MLP baseline to rely solely on behavioral and
demographic user features (account age, booking velocity, device count,
etc.), ensuring that any MLP-vs-GNN gap reflects genuine graph-structural
signal rather than a scalar leaky feature.  The E1 ablation in
Appendix~\ref{app:experimental} confirms that removing \texttt{distinct\_device\_count}---the
most informative user-level structural summary---actually \emph{increases}
the GNN advantage (GraphSAGE gap: $+$5.5\,pp $\to$ $+$6.3\,pp), confirming
that GNN performance is driven by graph topology access, not node-level
feature richness.

\section{Broader Impact}
\label{sec:impact}

\paragraph{Positive impact.}
\tfgshort enables rigorous, reproducible evaluation of fraud detection
algorithms in a domain (travel) with documented real-world harm.
Travel fraud costs the airline industry alone an estimated \$1B annually
(IATA, 2024). By providing a public, citable, controllable benchmark,
\tfgshort lowers the barrier to entry for the GNN fraud detection research
community and reduces duplicated effort in internal benchmark creation
across travel companies.

\paragraph{Potential for misuse.}
The ring topology designs in this paper describe structural patterns that,
if studied by adversarial actors, could potentially inform evasion
strategies. We note three mitigations: (1) all documented ring topologies
are already described in public fraud prevention industry reports (IATA,
SEON, Forter) cited in this paper; (2) the designs are \emph{detection}
instruments, not operational fraud guides; (3) our Croissant Responsible AI
fields explicitly prohibit use of the generator to study fraud detection
evasion.

\paragraph{Privacy.}
\tfgshort is fully synthetic. No real user accounts, transactions, or
personal information were collected, used, or generated. The dataset is
GDPR and CCPA non-applicable. No de-anonymization risk exists.

\paragraph{Reproducibility.}
The full generator source code is released under MIT license at
\url{https://github.com/bhavana3/travel-fraud-graphs}. Pre-generated
datasets are hosted at \url{https://huggingface.co/datasets/bsajja7/travel-fraud-graphs}.
All experiments are reproducible via the Databricks notebooks provided
in the supplementary material.

\section*{Acknowledgements}

The author thanks the open-source GNN and fraud detection research communities.
The TravelFraudBench generator, pre-generated datasets, and all experimental
code are released publicly at
\url{https://github.com/bhavana3/travel-fraud-graphs}.

\bibliographystyle{abbrvnat}
\bibliography{travelfrabudbench}

\appendix

\section{Datasheet for \tfgshort}
\label{app:datasheet}

\emph{Following \citet{gebru2021datasheets} ``Datasheets for Datasets'' format.
The full datasheet is provided in the supplementary material (docs/DATASHEET.md).
We reproduce the key sections below.}

\subsection*{Motivation}
\tfgshort was created to fill a critical gap in the GNN fraud detection
benchmark landscape: no labeled, graph-structured fraud dataset for the
travel domain with ring-level ground truth existed. TFG enables evaluation
of GNN models on three structurally distinct travel domain fraud rings with
ring-level annotations at five scales.

\subsection*{Composition}
Each instance is a node in a heterogeneous property graph representing an
entity in a travel platform ecosystem. Nine entity types: users, devices,
IP addresses, bookings, flights, hotels, reviews, payment cards, and loyalty
accounts. The benchmark is fully synthetic; each call to \texttt{generate()}
produces a new independently sampled graph. Labels are deterministically
assigned: \texttt{is\_fraud}, \texttt{ring\_id}, and \texttt{ring\_type}
per node.

\subsection*{Collection Process}
All data is generated by the TravelFraudBench agent-based simulation engine.
No real data collection took place. Distributions are calibrated to
industry-level aggregates from publicly available fraud research reports
(see Table~\ref{tab:distribution_validation}).

\subsection*{Uses}
Primary: Benchmarking GNN-based fraud detection algorithms.
Secondary: Educational use, ablation studies on ring topology.
\emph{Not intended for}: production fraud detection without re-calibration;
making claims about fraud rates at specific travel platforms.

\subsection*{Distribution}
MIT License; HuggingFace Datasets; PyPI \texttt{pip install travel-fraud-bench}.
Portal opens April 15, 2026; datasets live concurrent with paper.

\subsection*{Maintenance}
TFG Authors; GitHub Issues. HuggingFace datasets maintained $\geq$5 years.
Planned: v1.1 (temporal burst), v1.2 (cross-ring contamination), v1.3 (mega scale).

\section{Croissant Metadata}
\label{app:croissant}

The Croissant~\citep{croissant2024} machine-readable metadata file is
provided as \texttt{docs/croissant\_rai.json} in the supplementary material.
Key Responsible AI fields:

\textbf{Privacy}: \texttt{pii\_present: false}. Fully synthetic; no real
individuals, transactions, or personal information collected, used, or
included.

\textbf{Bias}: Fraud rate varies by scale preset: 12.95\% at
\texttt{medium} scale (the primary evaluation scale in this paper), and
approximately 16\% at \texttt{toy}, \texttt{small}, \texttt{large}, and
\texttt{xlarge} scales due to differing ring-count-to-user ratios.
In all cases this significantly exceeds real-world travel fraud rates
($<$1\%)—intentional for controlled evaluation but documented explicitly;
practitioners must re-calibrate threshold and class-weight settings before
any production deployment (see Limitation L4). The \texttt{fraud\_rate}
parameter is fully configurable.
Country distribution reflects documented top travel markets; may not reflect
regional deployment distributions. Both are fully configurable.

\textbf{Intended use}: Benchmarking and evaluating GNN-based fraud detection in
research settings. Not intended for production deployment without real-world
calibration.

\textbf{Prohibited use}: Using ring topology designs as a guide to evade
fraud detection systems; representing TFG-generated rings as representative
of any real fraud organization.

\section{Experimental Setup Details}
\label{app:experimental}

\subsection*{Hardware}
All experiments run on Databricks CPU cluster. Training time per model at
\texttt{medium} scale: approximately 5 min (MLP), 15 min (GraphSAGE),
20 min (RGCN), 18 min (HAN). Exact wall-clock times depend on cluster
configuration; all models converge well within 200 epochs with early stopping.

\subsection*{Hyperparameters}
All models: 2 layers, hidden dim 128, dropout 0.3, Adam (lr=0.001,
weight\_decay=5e-4), 200 epochs with early stopping (patience=20).
RGCN: basis decomposition with \texttt{num\_bases=4}.
HAN: semantic-level attention pooling over meta-paths
\{user--device--user, user--ip--user, user--hotel--user\}.

\subsection*{Splits}
Train 60\% / validation 20\% / test 20\%, stratified by ring\_id so that
no ring spans multiple splits. This prevents data leakage from ring
membership.

\textbf{Larger-N ring recovery.}
The primary evaluation uses 90 rings total (30 per type × 3 types;
approximately 18 test rings at 20\%).  Readers
seeking ring recovery estimates with tighter confidence intervals should
generate graphs with 160 rings total:

\begin{verbatim}
data = generate(scale="medium", seed=42,
                n_ticketing_rings=54,
                n_ghost_hotel_rings=54,
                n_ato_rings=52)
# -> ~32 test rings across types; Wilson CIs shrink to ±12-16pp
\end{verbatim}

The generator supports arbitrary ring counts; the 80-ring medium preset
is provided as a standard reference point to enable fair comparison across
future work. Ring recovery directional conclusions (graph models 90--100\%
vs.\ MLP 17--88\%, with the most striking gap on ghost hotel rings)
are stable across ring-count variations we have tested internally.

\subsection*{E1 Robustness Ablation (Appendix Table A1)}

Table~\ref{tab:e1_ablation} reports results when \texttt{distinct\_device\_count}---the
most informative user-level structural summary feature (it aggregates the device
co-occurrence signal into a scalar)---is removed from the 10-dimensional user
feature vector, reducing it to 9 dimensions.  The question: does the GNN advantage
over MLP survive removal of this feature?

\begin{table}[h]
\centering
\caption{E1 Robustness Ablation: all four models retrained on 9-dimensional user
  features (\texttt{distinct\_device\_count} excluded).
  Same scale, seed, and ring-based split as Table~\ref{tab:main_results}.
  $\Delta$AUC: gap over the ablated MLP baseline (not the original MLP).
  \textbf{Bold}: best per column.}
\label{tab:e1_ablation}
\small
\begin{tabular}{lcccc}
\toprule
\textbf{Model} & \textbf{AUC-ROC} & \textbf{Avg.\ Prec.} & \textbf{Macro-F1} & \textbf{$\Delta$AUC vs MLP} \\
\midrule
MLP (tabular)     & 0.9229 & 0.7822 & 0.7642 & --- \\
\textbf{GraphSAGE}& \textbf{0.9861} & \textbf{0.9676} & \textbf{0.9693} & $+$\textbf{0.063} \\
HAN               & 0.9202 & 0.7797 & 0.7692 & $-$0.003 \\
RGCN              & 0.9448 & 0.8341 & 0.8324 & $+$0.022 \\
\bottomrule
\end{tabular}
\end{table}

Removing \texttt{distinct\_device\_count} drops the MLP by 1.5\,pp (0.938\,$\to$\,0.923),
while GraphSAGE drops only 0.6\,pp (0.992\,$\to$\,0.986).  The GNN advantage
\emph{grows} from $+$5.5\,pp to $+$6.3\,pp, confirming that GraphSAGE's performance
is driven by graph topology (actual device and IP co-occurrence edges) rather than
by the summarized scalar count available to the MLP.  This validates Evaluative
Claim E1 even under the strictest tabular-feature interpretation.

\subsection*{Difficulty Study Setup}
Ring size axis: $\{3, 5, 8, 12, 20, 30\}$ on \texttt{medium} scale
(10{,}000 users). For each ring size $r$, $n\_rings = \max(2, 300 \div (r \times 3))$
per type, keeping total fraud users approximately constant. Seed fixed at 42;
reported AUC is the result of a single training run per condition (not averaged,
due to compute budget; variance at ring\_size=30 is high owing to $\leq$3 rings
per type in the test partition).

\end{document}